\pdfoutput=1
\RequirePackage{fix-cm}
\documentclass{svjour3}                     % onecolumn (standard format)
\smartqed  % flush right qed marks, e.g. at end of proof
\usepackage{graphicx,amsmath,amssymb,amsfonts, cite}
\usepackage[round]{natbib}
\usepackage{hyperref}
\usepackage{subcaption}
\usepackage{mathptmx}      % use Times fonts if available on your TeX system
\begin{document}

\title{Uncertainty Estimation in Deep Neural Networks for Point Cloud Segmentation in Factory Planning}


\titlerunning{Uncertainty Estimation in Neural Networks for Point Cloud Segmentation in Factory Planning}        % if too long for running head

\author{Christina Petschnigg  \and
	J\"urgen Pilz %etc.
}


\institute{C. Petschnigg \at
              BMW Group, Munich, Germany \\
              \email{Christina.Petschnigg@bmw.de}           %  \\
	\and
	 J. Pilz \at 
	Alpen-Adria Universit\"at, Klagenfurt am W\"orthersee, Austria \\
	\email{Juergen.Pilz@aau.at}
}

\date{Received: date / Accepted: date}

\maketitle

\begin{abstract}
The digital factory provides undoubtedly a great potential for future production systems in terms of efficiency and effectivity. A key aspect on the way to realize the digital copy of a real factory is the understanding of complex indoor environments on the basis of 3D data. In order to generate an accurate factory model including the major components, i.e. building parts, product assets and process details, the 3D data collected during digitalization can be processed with advanced methods of deep learning. In this work, we propose a fully Bayesian and an approximate Bayesian neural network for point cloud segmentation. This allows us to analyze how different ways of estimating uncertainty in these networks improve segmentation results on raw 3D point clouds. We achieve superior model performance for both, the Bayesian and the approximate Bayesian model compared to the frequentist one. This performance difference becomes even more striking when incorporating the networks' uncertainty in their predictions. For evaluation we use the scientific data set S3DIS as well as a data set, which was collected by the authors at a German automotive production plant. The methods proposed in this work lead to more accurate segmentation results and the incorporation of uncertainty information makes this approach especially applicable to safety critical applications.
\keywords{Point Clouds \and 3D Segmentation \and Bayesian Deep Learning \and Dropout Training \and Uncertainty Estimation \and Digital Factory \and Factory Planning}
\end{abstract}

\section{Introduction}
A 3D model of factory buildings and inventory as well as the simulation of process steps play a major role in different planning domains. In general, virtual planning has many advantages compared to analogue planning. The most stringent benefit is the detection of planning mistakes early on in the planning process. This is favourable as planning mistakes are detected well before implementation~\citep{kuhn2006digital}, i.e. before factory ramp-up, before new machinery is ordered, before construction is under way or before the production process is detailed. This is due to the fact that the structure and layout of the building influences several other domains. A changing building model can entail changes in spatial availability for production or logistics assets. Thus, the layout of production lines or the concept of machines may have to be adapted accordingly. Further, virtual planning reduces travel efforts as planners do not have to meet on-site to discuss modifications or reorganizations. They can rather meet in a multi-user simulation model or a virtual reality supported 3D environment, which saves a substantial amount of travel time and cost. Digital 3D models are the basis for building reorganizations as well as the introduction of completely new or modified manufacturing process steps.\\
In order to determine the as-is state of a production plant there are several challenges to tackle. First of all, current data in the respective plant have to be collected. In order to acquire 3D information, laser scanning and photogrammetry are useful digitalization techniques. After plant digitalization the collected data have to be pre-processed including data cleaning and fusion of inputs from different sources. Working solely on the basis of point clouds for the sake of factory or process simulation is not possible as point clouds generated by laser scanners and photogrammetry techniques suffer from occlusions, which results in holes within the point cloud. For instance, the outcomes of collision checking are not reliable when the point cloud is not complete. Additionally, a point cloud does not contain any information on how to separate different objects. Therefore, the introduction of new or the displacement of existing objects is time consuming, as the respective set of points has to be selected manually. In order to separate different objects from one another automatically, a segmentation step has to be introduced~\citep{petschnigg2020point}.\\
Most of the existing deep learning architectures make use of the frequentist notion of probability. However, these so-called frequentist neural networks suffer from two major drawbacks. They do not quantify uncertainty in their predictions. Often, the softmax output of frequentist neural networks is interpreted as network uncertainty, which is, however, not a good measure. The softmax value only normalizes an input vector but cannot as such be interpreted as network (un)certainty~\citep{gal2016dropout}. Especially for out of distribution samples the softmax output can give rise to misleading interpretations~\citep{sensoy2018evidential}. In the case of deep learning frameworks being integrated into safety critical applications like autonomous driving it is important to know what the network is uncertain about. There was one infamous accident caused by a partly autonomous driving car that confused the white trailer of a lorry with the sunlit sky or a bright overhead sign~\citep{banks2018driver}. By considering network uncertainties similar scenarios could be mitigated. Another shortcoming of frequentist neural networks is their tendency to overfit on small data sets with a high number of features. In this work, however, we focus on uncertainty estimation rather than the challenge of feature selection.\\
We present a novel Bayesian 3D point cloud segmentation framework based on PointNet~\citep{qi2017pointnet} that is able to capture uncertainty in network predictions. The network is trained using variational inference with multivariate Gaussians with a diagonal covariance matrix as variational distribution. This approach adds hardly any additional parameters to be optimized during each backward pass~\citep{posch2019variational}. Further, we formulate an approximate Bayesian neural network by applying dropout training as suggested in~\citep{gal2016dropout}. Further, we use an entropy based interpretation of uncertainty in the network outputs and distinguish between overall, data related and model related uncertainty. These types of uncertainty are called predictive, aleatoric and epistemic uncertainty, respectively~\citep{chen2017multi}. It makes sense to consider this differentiation as it shows, which predictions are uncertain and to what extent this uncertainty can be reduced by further model refinement. The remaining uncertainty after model optimization and training is then inherent to the underlying data set. Other notions of uncertainty based on the variance or credible intervals of the predictive network outputs are discussed and evaluated. To the best of our knowledge no other work has treated the topic of uncertainty estimation and Bayesian training of 3D segmentation networks that operate on raw and unordered point clouds without a previous transformation into a regular format. Aside from an automotive data set that is collected by the authors at a German automotive manufacturing plant, the proposed networks are evaluated on a scientific data set in order to ensure the comparability with other state-of-the-art frameworks. Summing up, the contributions of this paper are:
\begin{itemize}
\item Workflow: We describe how to quantify uncertainty in segmentation frameworks that operate on raw and unstructured point clouds. Further, it is discussed how this information assists in generating current factory models.
\item Framework:  We formulate a PointNet~\citep{qi2017pointnet} based 3D segmentation model that is trained in a fully Bayesian way using variational inference and an approximate Bayesian model, which is derived by the application of dropout training.
\item Experiment: We evaluate how the different sources of uncertainty affect the neural networks' segmentation performance in terms of accuracy. Further, we outline how the factory model can be improved by considering uncertainty information.
\end{itemize}
The remainder of this paper is organized in the following way. Section~\ref{sec:literature-review} conducts a thorough literature review on 3D point cloud processing framework including deep neural networks, Bayesian neural networks and uncertainty quantification. In the subsequent Section~\ref{sec:model-description} the frequentist, the approximate Bayesian and the fully Bayesian models are described in more detail. Section~\ref{sec:data-sets} discusses the scientific and industrial data sets that are used for the evaluation of our models and elaborates their characteristics. The models are evaluated with respect to their performance in Section~\ref{sec:results-and-analysis}. Finally, Section~\ref{sec:discussion-and-conclusion} provides a discussion, which describes the bigger scope of this work and concludes the paper.

\label{sec:introduction}

\section{Literature Review}
\label{sec:literature-review}
The following paragraphs cast light upon prior research in the areas of segmentation of 3D point clouds as well as Bayesian neural networks and uncertainty estimation. The neural networks discussed in the first section are all based on the classical or frequentist interpretation of probability. Bayesian neural networks rather take on the Bayesian interpretation of probability, which views probability as a personal degree of belief. 

\subsection{3D Segmentation}
In contrast to images that have a regular pixel structure, point clouds are irregular and unordered. Further, they do not have a homogeneous point density due to occlusions and reflections. Neural networks that process 3D point clouds have to tackle all of these challenges. Most networks are based on the frequentist interpretation of probability and are divided into three classes based on the format of their input data. There are deep learning frameworks that consume voxelized point clouds~\citep{qi2016volumetric,wu20153d,zhou2018voxelnet,lei2020seggcn}, collections of 2D images derived by transforming 3D point clouds to the 2D space from different views~\citep{chen2017multi,feng2018towards,yang2018pixor} and raw unordered point clouds~\citep{qi2017pointnet,qi2017pointnet++,ravanbakhsh2016deep}. On the one hand, voxelization of point clouds has the advantage of providing a regular structure apt for the application of 3D convolutions. On the other hand, it renders the data unnecessarily big as unoccupied areas of the point cloud are still represented by voxels. Generally, this format conversion introduces truncation errors~\citep{qi2017pointnet}. Further, voxelization reduces the resolution of the point cloud in dense areas, leading to a loss of information~\citep{xie2020review}. Transforming 3D point clouds to 2D images from different views allows the application of standard 2D convolutions having the advantage of elaborate kernel optimizations. Yet, the transformation to a lower space can cause the loss of structural information embedded in the higher dimensional space. Additionally, in complex scenes a high number of viewports have to be taken into account in order to describe the details of the environment~\citep{xie2020review}. For this reason, the following work focuses on the segmentation of raw point clouds.\\
In order to generate a factory model out of raw point clouds, the objects of interest have to be detected and their pose needs to be estimated. One approach that extracts 6 degrees-of-freedom~(DoF) object poses, i.e. the translation and orientation with respect to a predefined zero point, in order to generate a simulation scene is presented in~\citep{avetisyan2019scan2cad}. The framework is called Scan2CAD and describes a frequentist deep neural network that consumes voxelized point clouds as well as computer-aided design~(CAD) models of eight household objects and directly learns the 6DoF CAD model alignment within the point cloud. The system presented in~\citep{avetisyan2019end} has similar input data and estimates the 9DoF pose, i.e. translation, rotation and scale, of the same household objects. A framework for the alignment of CAD models, which is based on global descriptors computed by using the Viewpoint Feature Histogram approach~\citep{rusu2010fast} rather than neural networks, is discussed in~\citep{aldoma2011cad}. Generally, direct 6DoF or 9DoF pose estimation on the basis of point clouds and CAD models can be used to set up environment models and simulation scenes. However, these approaches always require the availability of CAD models, which is not the case for many building and inventory objects in real-world factories. Thus, we follow the approach of semantic segmentation instead of direct pose estimation. Semantic segmentation allows us to extract reference point clouds of objects, for which no CAD model is available. These objects can either be modelled in CAD automatically by using meshing techniques or by hand if the geometry is too difficult to capture realistically. Further, the segmentation approach enables us to part the point cloud into bigger contexts, i.e. subsets of points belonging to the construction, assembly or logistics domain. These smaller subsets of points can be sent to the respective departments for further processing, reducing the computational burden of the point cloud to be processed. Mere pose estimation is not sufficient to fulfil this task. Aside from the semantic segmentation of point clouds, this work focuses on the formulation of Bayesian neural networks and how to leverage the uncertainty information that can be calculated in order to increase the models' accuracy.

\subsection{Bayesian Deep Learning and Uncertainty Quantification}
\label{subsec:literature-bayesian-dl}
In contrast to frequentist neural networks, where the network paramteters are point estimates, Bayesian neural networks (BNNs) place a distribution over each of the network parameters. For this reason, a prior distribution is defined for the parameters. After observing the training data the aim is to calculate the respective posterior distribution, which is difficult as it requires the solution of a generally intractable integral. There exist several solution approaches including variational inference~(VI)~\citep{blundell2015weight,graves2011practical}, Markov Chain Monte Carlo~(MCMC) methods~\citep{brooks2011handbook,gelfand1990sampling,hastings1970monte}, Hamiltonian Monte Carlo (HMC) algorithms~\citep{duane1987hybrid} and Integrated Nested Laplace approximations~(INLA)~\citep{rue2009approximate}. VI provides a fast approximation to the posterior distribution. However, it comes without any guaranteed quality of approximation. MCMC methods in contrast are asymptotically correct but they are computationally much more expensive than VI. Even the generally faster HMC methods are clearly more time consuming than VI~\citep{blei2017variational}. As the data sets used for evaluating this work are huge in size, we decide to apply VI due to efficiency reasons.\\
In literature there are several ways of how uncertainty can be quantified in BNNs. It is possible to distinguish between data and model related uncertainty, which are referred to as aleatoric and epistemic uncertainty, respectively~\citep{der2009aleatory}. The overall uncertainty inherent to a prediction can be computed as the sum of aleatoric and epistemic uncertainty and is called predictive uncertainty. Such a distinction is beneficial for practical applications in order to determine to what extent model refinement can reduce predictive uncertainty and to what extent uncertainty stems from the data set itself. One possibility to describe predictive uncertainty $U_{pred}$ is based on entropy, i.e. $U_{pred} = \mathbb{H}[y^{\star}\vert \underline{w},\underline{x}^{\star}]$~\citep{gal2017deep}. Another way of quantifying uncertainty in the network parameters of BNNs is presented in~\citep{posch2019variational}. This approach introduces only two uncertainty parameters per network layer, which allows us to grasp uncertainty layer-wise but does not impair network convergence. The overall model uncertainty is measured by estimating credible intervals of the predictive network outputs $p(y^{\star}\vert \underline{w}^{k},\underline{x}^{\star})$. This is based on the notion that higher uncertainty in the network parameters results in higher uncertainty in the network outputs. Further, the predictive variance can be used for uncertainty estimation as well.

\section{Model Descriptions}
\label{sec:model-description}
In the following the frequentist, the approximate Bayesian and the fully Bayesian model are explained. In order to formulate these models let $X=\{\underline{x}_{1},\dots,\underline{x}_{n}\}$ be the input data and \mbox{$Y=\{y_{1},\dots,y_{n}\}$} the corresponding labels, where $y_{i}\in\{1,\dots,m\},~m\in\mathbb{N},~i\in\{1,\dots,n\},~n\in\mathbb{N}$.  Further, let $\mathcal{W}$ and $\mathcal{B}$ denote all the network parameters including the weights and biases, respectively.  The network weights and biases of the $i$-th network layer are denoted by $\mathcal{W}_{i}$ and $\mathcal{B}_{i},~i\in\{1,\dots,d\}$, where $d\in\mathbb{N}$ is the network depth. The described network architectures mainly apply convolutional layers, thus, we write $conv(i,j)$ for a convolutional layer with input dimension $i\in\mathbb{N}$ and output dimension $j\in\mathbb{N}$. In the following $\sigma(\cdot)$ denotes a non-linear function. In the sequel, uncertainty estimation is explained in more detail and its practical implementation is discussed.

\subsection{Frequentist PointNet}
The baseline for the following derivations and evaluations is the PointNet segmentation architecture~\citep{qi2017pointnet}. This framework consumes raw and unordered point sets in a block structure. The number of points in each of the blocks is exactly 4096 - either due to random down-sampling or due to up-sampling by repeated drawing of points. Each input point is represented by a vector $\underline{x}$ containing xyz-coordinates centred about the origin and RGB values. For later illustration purposes, we add another three dimensions, which hold the original point coordinates, i.e. $dim(\underline{x})=9$. The actual network input is a tensor of dimension $bs \times 4096 \times 6$, where $bs\in\mathbb{N}$ represents the batch size. The batch size corresponds to the number of input blocks being treated at a time. Each of these blocks consists of exactly $4096$ points. Further, the centred point coordinates are rather used for network training than the original ones, thus, the last dimension is $6$ instead of $9$.\\
In this architecture a symmetric input transformation network is applied first. It is followed by a convolutional layer $conv(6,64)$ and a feature transformation network. After the feature transformation another two convolutional layers $conv(64,128,1024)$ are applied before extracting global point cloud features using a max pooling layer. These global features are concatenated to the local features, which correspond to the direct output of the feature transformation network. The resulting network scores are generated by four convolutional layers $conv(1088,512,256,128,m)$, where $m\in\mathbb{N}$ is the number of classes. The rectified linear unit (ReLU) is used as a non-linear activation function in this network.

\subsection{Approximate Bayesian PointNet}
For the approximate Bayesian PointNet segmentation network we use the notion that dropout training in neural networks corresponds to approximate Bayesian inference~\citep{gal2016dropout}. In the following, this network will be referred to as dropout PointNet. Dropout in a single hidden layer neural network can be defined by sampling binary vectors $\underline{c}_{1}\in\{0,1\}^{d_{1}}$ and $\underline{c}_{2}\in\{0,1\}^{d_{2}}$ from a Bernoulli distribution such that $c_{1,q}\sim Be(p_{1})$ and $c_{2,k}\sim Be(p_{2})$, where $q=1,\dots,d_{1}$ and $k=1,\dots,d_{2}$. The variables $d_{1}$ and $d_{2}$ corresponds to the number of weights in the respective layer and $p_{1},~p_{2}\in [0,1]$. Then dropout can be interpreted as
\begin{equation}
\underline{\hat{y}}~=~\sigma(\underline{x}(\underline{c}_{1}\mathcal{W}_{1})+\mathcal{B}_{1})(\underline{c}_{2}\mathcal{W}_{2}).
\end{equation}
The bias in the second layer is omitted, which corresponds to centring the output. The network output $\underline{\hat{y}}$ is normalized using the softmax function
\begin{equation}
\hat{p}_{ij}=\frac{exp(\hat{y}_{ij})}{\sum_{j'=1}^{m}exp(\hat{y}_{ij'})}, ~i=1,\dots,n,~j=1,\dots,m.
\end{equation}
The log of this function results in the log-softmax loss. In order to improve the generalization ability of the network, $\mathcal{L}_{2}$ regularization terms for the network weights and biases can be added to the loss function. The optimization of such a neural network acts as approximate Bayesian inference in deep Gaussian process models~\citep{gal2016dropout}. This approach neither changes the model nor the optimization procedure, i.e. the computational complexity during network training does not increase. It is suggested to apply dropout before every weight layer in the network, however, empirical results with respect to convolutional neural networks show inferior performance when doing so. Thus, we place dropout before the last three layers in the PointNet model, with a dropout probability of $0.1$. Other than that the frequentist model is left unchanged. Placing dropout within the input or feature transform network results in considerably lower performance.

\subsection{Bayesian PointNet}
As already mentioned, BNNs place a distribution over each of the network parameters. In Bayesian deep learning all of the network parameters including weights and biases are expressed as one single random vector $\underline{\mathcal{W}}$.  The prior knowledge about these parameters is captured by the a priori distribution $p(\underline{w})$. After observing some data $(X,Y)$ the a posteriori distribution can be derived. Using Bayes' Theorem the posterior density reads
\begin{equation}
p(\underline{w}\vert Y,X)~=~\frac{p(Y\vert \underline{w},X)p(\underline{w})}{\int p(Y\vert \underline{w},X)p(\underline{w})d\underline{w}}.
\end{equation}
The likelihood $p(Y\vert \underline{w},X)$ is given by $\prod_{i=1}^{n}BNN(\underline{x}_{i};\underline{w})_{y_{i}}$, which corresponds to the product of the BNN outputs for all training inputs under the assumption of stochastic independence. However, the integral in the denominator is usually intractable, which makes the direct computation of the posterior difficult. In Section~\ref{subsec:literature-bayesian-dl} different methods for posterior approximation are discussed. As already mentioned, we use VI as it is most efficient in the case of a huge amount of training data. The idea of VI is to approximate the posterior $p(\underline{w}\vert Y,X)$ by a parametric distribution $q_{\underline{\varphi}}(\underline{w})$. To this end the Kullback-Leibler divergence (KL-divergence) between the variational and posterior density is minimized, i.e.
\begin{equation}
KL(q_{\underline{\varphi}}(\underline{w})\| p(\underline{w}\vert Y,X))~:=~\mathbb{E}_{q_{\underline{\varphi}}(\underline{w})}\left(ln\frac{q_{\underline{\varphi}}(\underline{w})}{p(\underline{w}\vert Y,X)}\right)~=~\int q_{\underline{\varphi}}(\underline{w})~ln\frac{q_{\underline{\varphi}}(\underline{w})}{p(\underline{w}\vert Y,X)}d\underline{w}.
\end{equation}
The KL-divergence does not describe a real distance metric as the triangle inequality and the property of symmetry are not fulfilled. Nevertheless it is frequently used in BNN literature to measure the distance between two distributions. Due to the unknown posterior in the denominator of the KL-divergence, it cannot be optimized directly. According to~\citep{bishop2006pattern} the minimization of the KL-divergence is equivalent to the minimization of the log evidence lower bound (ELBO), which reads
\begin{equation}
ELBO ~=~ -\int q_{\underline{\varphi}}(\underline{w}) ln p(Y\vert \underline{w},X)d\underline{w}~+~KL(q_{\underline{\varphi}}(\underline{w})\| p(\underline{w})).
\end{equation}
After the optimization of the variational distribution it can be used to approximate the posterior predictive distribution for unseen data. Let $\underline{x}^{\star}$ be an unseen input with corresponding label $y^{\star}$. The posterior predictive distribution represents the belief in a label $y^{\star}$ for an input $\underline{x}^{\star}$ and is given by
\begin{equation}
p(y^{\star}\vert \underline{x}^{\star},Y,X)~=~\int p(y^{\star}\vert \underline{w},\underline{x}^{\star})p(\underline{w}\vert Y,X)d\underline{w}.
\end{equation}
The two factors under the integral correspond to the (future) likelihood and the posterior. The intractable integral can be approximated by Monte Carlo integration using $K\in\mathbb{N}$ terms and the posterior distribution is replaced by the variational distribution, i.e. 
\begin{equation}
p(y^{\star}\vert \underline{x}^{\star},Y,X)~\approx~\frac{1}{K}\sum_{k=1}^{K} BNN(\underline{x}^{\star}; \underline{\hat{w}}_{k})_{y^{\star}}~~~\mbox{with}~\underline{\hat{w}}_{k}\underset{i.i.d.}{\sim}q_{\underline{\varphi}}(\underline{w}),
\end{equation}
with $BNN$ denoting a forward pass through the network and $\underline{\hat{w}}_{k}$ is the $k$-th weight sample. Finally, the prediction $\hat{y}^{\star}$ is given by the index of the largest element in the mean of the posterior predictive distribution and thus reads
\begin{equation}
\hat{y}^{\star} ~ = ~ \underset{j\in\{1,\dots,m\}}{\mbox{arg max}}~\frac{1}{K}\sum_{k=1}^{K} BNN(\underline{x}^{\star}; \underline{\hat{w}}_{k})_{j}.
\end{equation}
After having discussed the theoretical background, we describe our Bayesian model and the corresponding variational distribution. The model we suggest has a similar structure to the framework in~\citep{posch2019variational}. The weights $\mathcal{W}_{i}$ and biases $\mathcal{B}_{i}$ of the $i$-th network layer $i\in\{1,\dots,d\}$ are defined as follows
\begin{eqnarray}
\tau_{wi} &:=& log(1+exp(\delta_{wi}))\\
\tau_{bi} &:=& log(1+exp(\delta_{bi}))\\
\mathcal{W}_{i} &:=& \underline{\mu}_{wi}\odot (\underline{1}_{d_{i}}+\tau_{wi}\underline{\varepsilon}_{wi})\\
\mathcal{B}_{i} &:=& \underline{\mu}_{bi}\odot (\underline{1}_{d_{i}^{'}}+\tau_{bi}\underline{\varepsilon}_{bi}),
\end{eqnarray}
where $\delta_{wi}\in\mathbb{R},~\delta_{bi}\in\mathbb{R},~\underline{\mu}_{wi}\in\mathbb{R}^{d_{i}}$ and $\underline{\mu}_{bi}\in\mathbb{R}^{d_{i}^{'}}$ are the variational parameters. Further, $\underline{1}_{d}$ denotes the $d$-dimensional vector consisting of all ones, $\underline{\varepsilon}_{wi}\in\mathbb{R}^{d_{i}}$ as well as $\underline{\varepsilon}_{bi}\in\mathbb{R}^{d_{i}^{'}}$ are multivariate standard normally distributed and $\odot$ represents the Hadamard product. Thus, the weights and biases follow a multivariate normal distribution with a diagonal covariance matrix, i.e.
\begin{eqnarray}
\mathcal{W}_{i} & \sim & \mathcal{N}(\underline{\mu}_{wi},~\tau_{wi}diag(\underline{\mu}_{wi})^{2})\\
\mathcal{B}_{i} & \sim & \mathcal{N}(\underline{\mu}_{bi},~\tau_{bi}diag(\underline{\mu}_{bi})^{2})
\end{eqnarray}
For more detailed insights on the respective gradient updates see~\citep{posch2019variational}. Due to the dying ReLU problem, we use the leaky ReLU activation function in the Bayesian model with a negative slope of 0.01. The mean of the weights is initialized using the Kaiming normal initialization with the same negative slope as for the leaky ReLU activation.

\subsection{Uncertainty Estimation}
As already described the estimated uncertainty can be split into predictive, aleatoric and epistemic uncertainty. In practice, predictive uncertainty $U_{pred}$ is approximated by marginalization over the weights,
\begin{equation}
U_{pred}~\approx~-\sum_{y^{\star}\in Y}\left(\frac{1}{K}\sum_{k=1}^{K}p(y^{\star}\vert \underline{\hat{w}}^{k},~\underline{x}^{\star})\right)\cdot log\left(\frac{1}{K}\sum_{k=1}^{K}p(y^{\star}\vert \underline{\hat{w}}^{k},~\underline{x}^{\star})\right).
\label{eq:U-pred}
\end{equation}
In Equation~(\ref{eq:U-pred}) $p(y^{\star}\vert \underline{w}^{k},~\underline{x}^{\star})$ corresponds to the predictive network output of label $y^{\star}$ for an input data point $\underline{x}^{\star}$ and the $k$-th weight sample $\underline{\hat{w}}^{k}$ of the variational distribution. The total number of Monte Carlo samples is given by $K\in\mathbb{N}$. Aleatoric uncertainty $U_{alea}$ is interpreted as the average entropy over all the weight samples,
\begin{equation}
U_{alea}~=~\mathbb{E}_{q_{\underline{\varphi}}(\underline{w})}[\mathbb{H}[y^{\star}\vert \underline{\hat{w}}^{k},~\underline{x}^{\star}]] ~\approx~ -\frac{1}{K}\sum_{k=1}^{K}\sum_{y^{\star}\in Y}p(y^{\star}\vert \underline{\hat{w}}^{k},~\underline{x}^{\star})\cdot log(p(y^{\star}\vert \underline{\hat{w}}^{k},~\underline{x}^{\star})).
\label{eq:U-alea}
\end{equation}
Finally, epistemic uncertainty $U_{ep}$ is the difference between predictive uncertainty and aleatoric uncertainty, i.e. \mbox{$U_{ep}=U_{pred}-U_{alea}$}. Further, uncertainty in network prediction can be quantified by calculating the variance for the predictive network outputs. Another way is to calculate a credible interval of network outputs for each class. For instance, the 95~\%-credible interval can be calculated for each class. In the case that the 95~\%-credible interval of the predicted class overlaps with the 95~\%-credible interval of any other class, the prediction is considered to be uncertain.

\section{Data Sets}
\label{sec:data-sets}
Two different data sets are used to evaluate our Bayesian and the approximate Bayesian segmentation approach, which forms the core contribution of this work. The first one is the Stanford large-scale 3D indoor spaces data set that is open to scientific use and thus ensures the comparability of our approach to other methods. The second data set is a large-scale point cloud data set collected and pre-processed by the authors at a German automotive OEM.

\subsection{Stanford Large-Scale 3D Indoor Spaces Data Set}
The Stanford large-scale 3D indoor spaces (S3DIS) data set~\citep{armeni20163d} is an RGB-D data set of 6 indoor areas. It features more than 215 million points collected over an area totalling more than 6 000 m$^{2}$. The areas are spread across three buildings including educational facilities, offices, sanitary facilities and hallways. The annotations are provided on instance level and they distinguish 6 structural elements from 7 furniture elements. This totals the 13 classes including the building structures of ceiling, floor, wall, beam, column, window and door as well as the furniture elements of table, chair, sofa, bookcase, board and clutter. The data set can be downloaded from http://buildingparser.stanford.edu/dataset.html.

\subsection{Automotive Factory Data Set}
\label{subsec:automotive-factory-dataset}
\begin{figure}
\begin{subfigure}[c]{0.5\textwidth}
\includegraphics[width=0.90\textwidth]{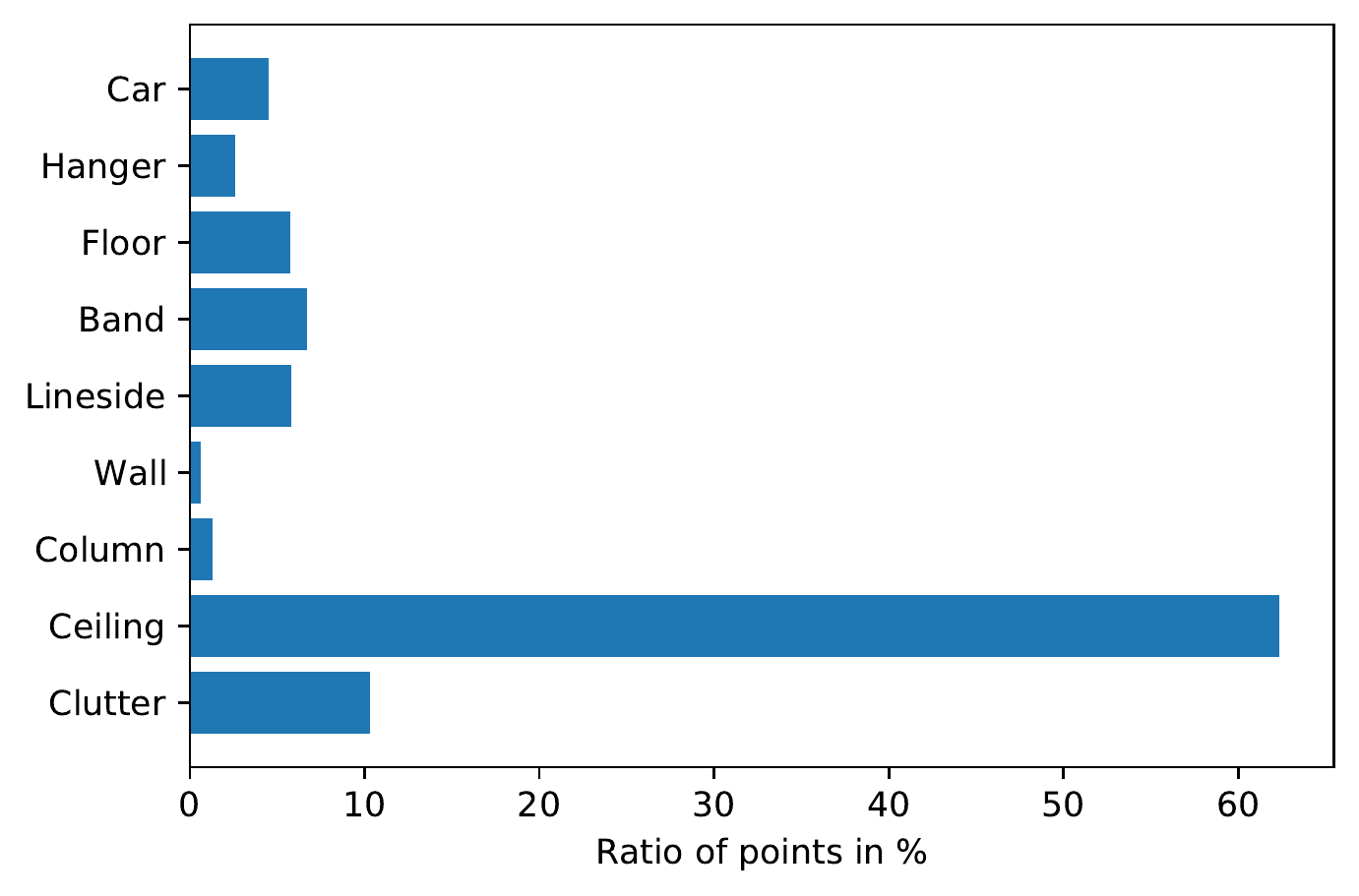}
\subcaption{}
\end{subfigure}
\begin{subfigure}[c]{0.5\textwidth}
\includegraphics[width=0.90\textwidth]{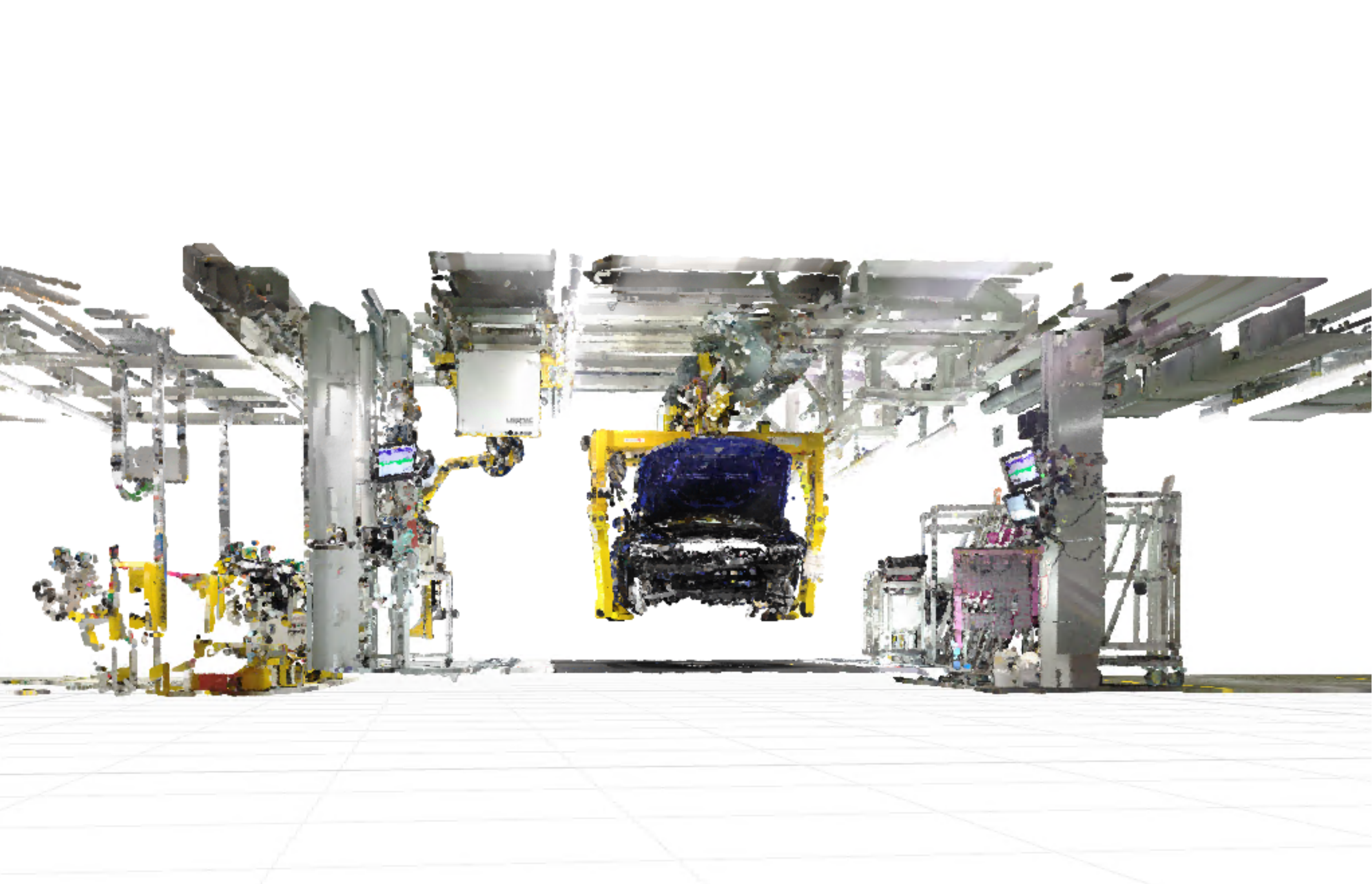}
\subcaption{}
\end{subfigure}
\caption{Overview of the collected data set. (\textbf{a}) Total ratio of all points in the point cloud that belong to each of the classes in \%. (\textbf{b}) Illustration of a point cloud displaying one tact of car body assembly.}
\label{fig:automotive-data-set}
\end{figure}

This data set was collected using both the static Faro Focus3D X 130HDR laser scanner and two DSLR cameras. In more detail a \mbox{Nikon D5500 camera} with an 8~mm fish-eye lens and a \mbox{Sony Alpha 7R II} with a 25~mm fixed focal length lens were used. We generate a global point cloud comprising 13 tacts of car body assembly by registration of several smaller point clouds collected at each scanner position. The final point cloud comprises more than one billion points before further pre-processing. The cleaning process is achieved using noise filters for coarse cleaning and fine tuning is done by hand. The resulting point set consists of 594~147~442 points. This accounts for a reduction of about 40~\% of the points after point cloud cleaning. Most of the removed points are noise points caused by reflections and the blur of moving objects like people walking by the laser scanner. The data set is divided into 9 different classes, namely car, hanger, floor, band, lineside, wall, column, ceiling and clutter. The labelling is done manually by the authors. The class clutter is a placeholder for all the objects that cannot be assigned to one of the other classes. All of the remaining classes are either building structures or objects that can only be moved with high efforts, thus, they are essentially immovable and have to be considered during planning tasks. The resulting data set is highly imbalanced with respect to the class distribution. Figure~\ref{fig:automotive-data-set}~(a) depicts the class distribution of this data set. Clearly, there is a notable excess of points belonging to the class ceiling and relatively few points belong to the classes of wall and column. This is mainly due to the layered architecture of the ceiling that results in points belonging to the structure on various heights. As walls and columns are mostly draped with other objects like tools, cables, fire extinguishers, posters and information signs, there is only a small number of points that truly belongs to the classes of wall and column. This is also the reason why especially these two classes suffer from a high degree of missing data, i.e. holes in the point cloud. Due to this inhomogeneous class distribution, any segmentation system has to cope with this class imbalance. Figure~\ref{fig:automotive-data-set}~(b) illustrates the point cloud of one tact of car body assembly.

\section{Results and Analysis}
\label{sec:results-and-analysis}
The proposed networks are evaluated on our custom automotive factory data set as well as the scientific data set S3DIS. The segmentation performance is measured with respect to their accuracy and the mean intersection over union. Further, the described ways of uncertainty quantification are evaluated in terms of accuracy after disregarding uncertain predictions. All the considered models are implemented using Python's open source library PyTorch~\citep{NEURIPS2019pytorch}. The input point clouds comprising rooms or assembly tacts are cut into blocks and the number of points within these blocks is sampled to $4096$. These blocks serve as input for all networks. All models are trained using mini-batch stochastic gradient descent with a batch size of $16$ on the S3DIS and the automotive factory data set for the frequentist and the proposed dopout and Bayesian networks. The momentum parameter is set to $0.9$ for all models. A decaying learning rate $lr$ is used with an initial learning rate of $lr = 0.001$ in the frequentist and the dropout model as well as $lr = 0.01$ in the Bayesian model. The learning rate is decayed every $10$ epochs by a factor of $0.7$ during frequentist and dropout training as well as $0.9$ during Bayesian training. The batch size and the learning rate are optimized by using grid search and cross-validation.  In the approximate Bayesian neural network dropout is applied before the last three convolutional layers and the dropout rate is set to $0.1$ for the automotive factory data set. In the case of the S3DIS data set dropout is only applied before the last convolutional layer with a dropout rate of $0.1$. As we do not have dedicated prior information for the Bayesian model, the prior indicates that the parameter values should not diverge. Thus, we choose a prior expectation of zero for all parameters and a standard deviation of $4$ and $8$ for all weights and biases, respectively. In terms of approximating the posterior predictive distribution, we draw $K = 50$ Monte Carlos samples. All the considered models converge and training is stopped after 100 epochs.

\subsection{Segmentation Accuracy}
The three architectures, i.e. the frequentist, dropout and Bayesian PointNet~(PN), described in Section~\ref{sec:model-description} are evaluated in the following. The evaluation metrics used are the accuracy and the mean Intersection over Union (IoU). The accuracy is calculated by the number of correctly classified points divided by the total number of points. The IoU or Jaccard coefficient describes the similarity between two sets with finite cardinality. It is defined by the number of points of the intersection divided by the number of points of the union of the two sets. In this case, we evaluate the overlap between the points classified as class $i$ by the model and the points of class $i$ in the ground truth. Thus, the IoU for class $i$ reads
\begin{equation}
J_{i}~=~\frac{\mbox{\# points correctly classified as $i$}}{\mbox{\# points classified as $i$}~+~\mbox{\# number points in class $i$ in ground truth}}~,
\end{equation}
where $i\in\{1,\dots,m\}$ indicates the class label. The mean IoU is calculated as the mean IoU value over all classes. Table~\ref{tab:class-vs-bayes} illustrates that Bayesian PointNet clearly surpasses the performance of frequentist and dropout PointNet with respect to accuracy as well as mean IoU on the test set of both data sets.
\begin{table}
\caption{Segmentation results of classical, dropout and Bayesian PointNet on S3DIS and our automotive factory data set. For every model the training converges and is carried out for 100 epochs.}
\centering
\begin{tabular}{ccccc}
\hline
\textbf{Model} & \textbf{Data} & \textbf{Training Acc.} & \textbf{Test Acc.} &  \textbf{Test mIoU}\\
\hline
Classical PointNet & S3DIS & 95.44 \% & 87.81 \% & 0.6977\\
Dropout PointNet & S3DIS &  95.42 \% &  87.71 \% & 0.6921\\
Bayesian PointNet & S3DIS & \textbf{95.52 \%} & \textbf{88.57 \%} & \textbf{0.7042}\\
\hline
Classical PointNet & Automotive & 97.66 \% & 94.23 \% & 0.7808\\
Dropout PointNet & Automotive &  98.01 \% &  94.59 \% & 0.7972\\
Bayesian PointNet & Automotive & \textbf{98.66 \%} & \textbf{95.47 \%} & \textbf{0.8263}\\
\hline
\end{tabular}
\label{tab:class-vs-bayes}
\end{table}
For the S3DIS data set we test the models on area 6 and for the automotive factory data set we set aside two distinct assembly tacts. The prior information in the Bayesian model acts as additional observations and is thus able to reduce overfitting and increase model performance. An even more striking difference in performance will be illustrated in the next section when the information provided by uncertainty estimation is considered.

\subsection{Uncertainty Estimation}
\begin{figure}
\includegraphics[width=1\textwidth]{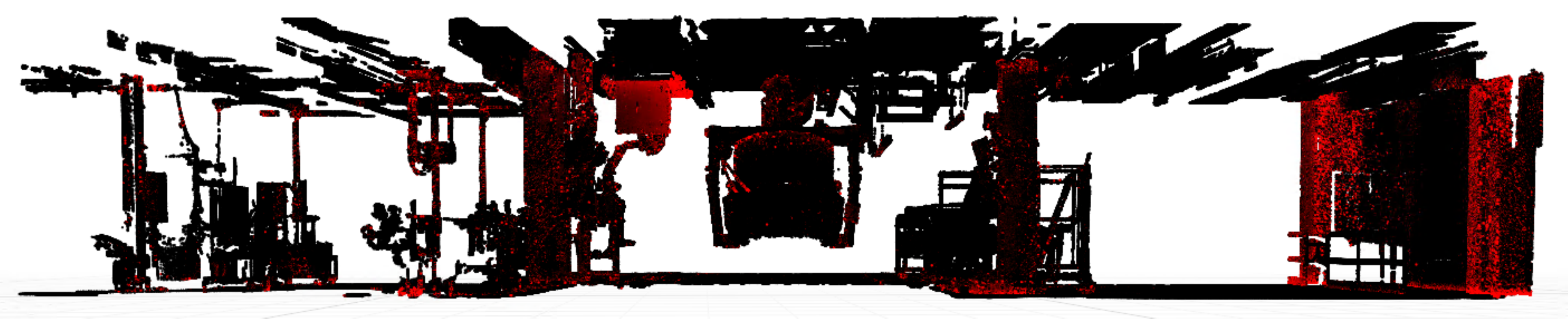}
\caption{Visualization of an assembly tact of the test set, where certain predictions are displayed in black and uncertain predictions are displayed in red. In this case the network (un)certainty is evaluated using the method based on credible interval estimation.}
\label{fig:assembly-tact-inc-unc}
\end{figure}

\begin{figure}
\begin{subfigure}[c]{0.5\textwidth}
\includegraphics[width=1\textwidth]{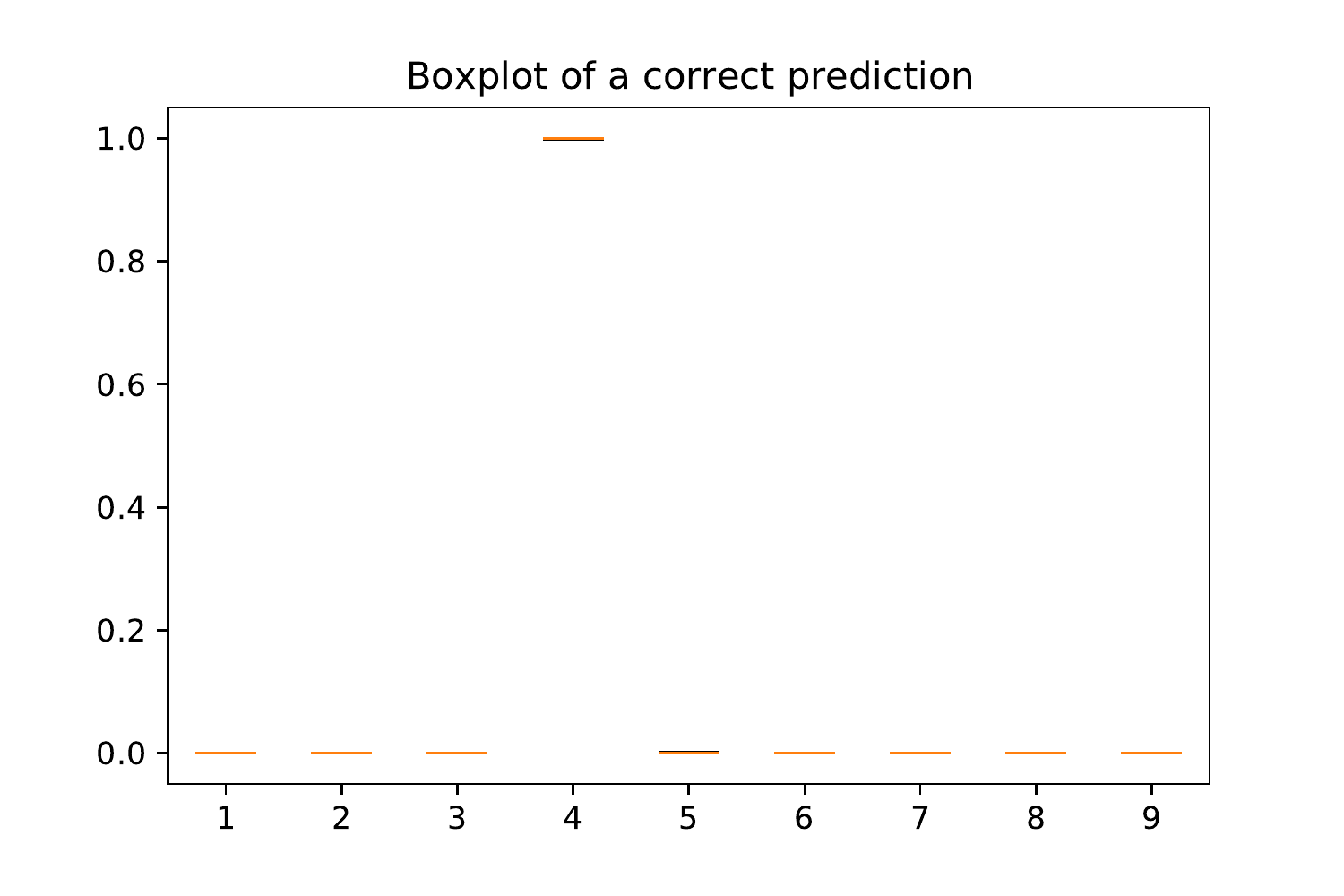}
\subcaption{}
\end{subfigure}
\begin{subfigure}[c]{0.5\textwidth}
\includegraphics[width=1\textwidth]{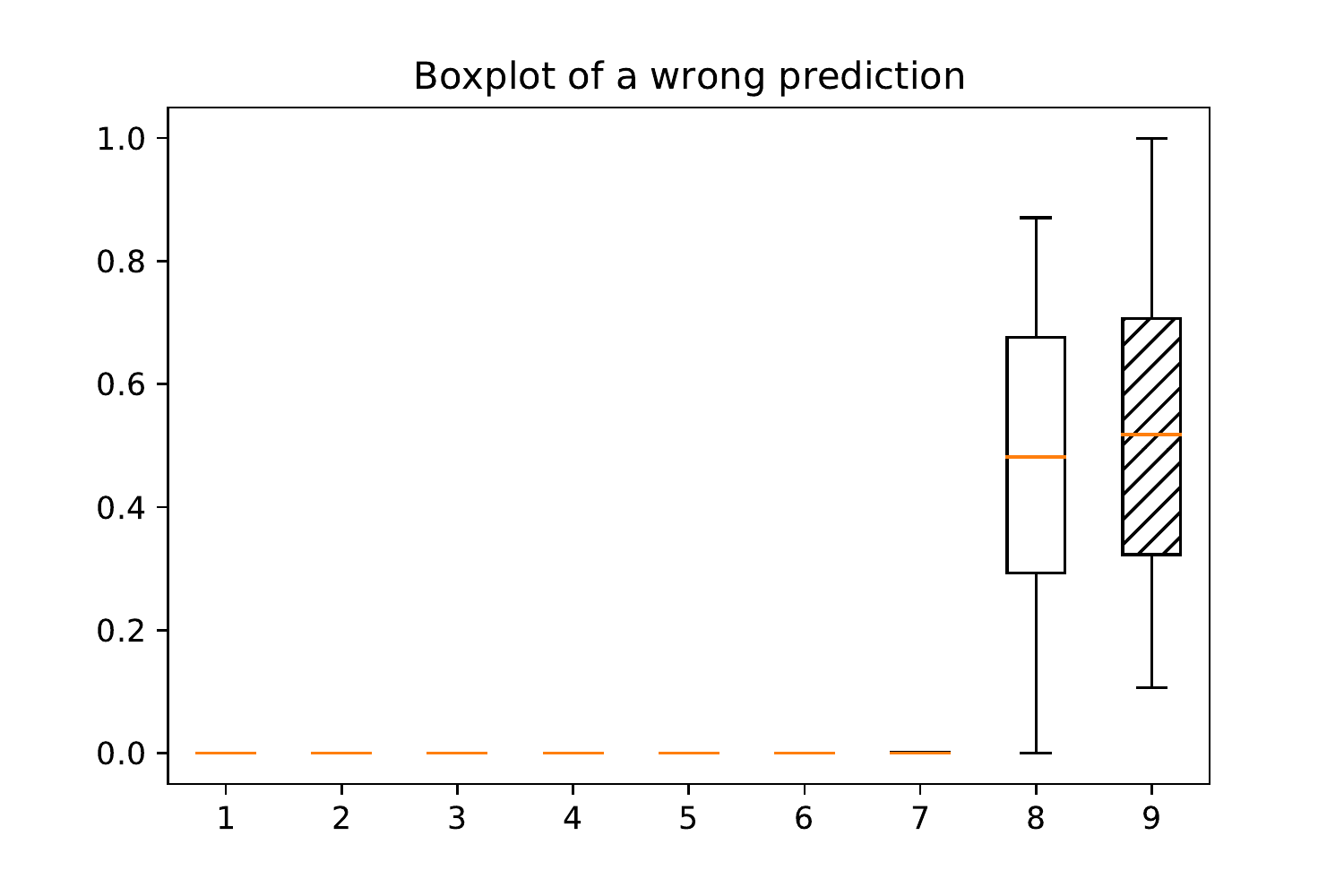}
\subcaption{}
\end{subfigure}
\caption{Boxplot of the predictive softmax outputs of Bayesian PointNet: (\textbf{a}) Boxplot of a correct prediction. The network is certain about its prediction as nearly all the propability mass is put on the correct class. (\textbf{b}) Boxplot of a wrong prediction. The network is uncertain about its prediction as the boxes of the true class and the predicted class overlap. The true class is represented by the white box and the wrongly predicted class is illustrated by the shaded box.}
\label{fig:boxplot}
\end{figure}

As already mentioned we estimate uncertainty using an entropy based approach, the predictive variance as well as an approach based on estimating credible intervals on the probabilistic network outputs. Predictive and aleatoric uncertainty are calculated as suggested in the Equations~(\ref{eq:U-pred}) and~(\ref{eq:U-alea}). Epistemic uncertainty is the difference of these two quantities. The predictive variance is determined on the basis of $K=50$ forward passes of the input through the network using the unbiased estimator for the variance. Based on the same sample the 95~\%-confidence intervals of the network outputs for each class are calculated. Table~\ref{tab:uncertainty-accuracy-bayes} contains the results of Bayesian PointNet for one room of each room type in area 6 of S3DIS data set as well as one tact of car body assembly belonging to the test data set. 
\begin{table}
\caption{Evaluation of different methods for uncertainty estimation and influence on model accuracy in the Bayesian model.}
\centering
%% \tablesize{} %% You can specify the fontsize here, e.g., \tablesize{\footnotesize}. If commented out \small will be used.
\begin{tabular}{ccccccc}
\hline
& \textbf{Bayesian PN} & \textbf{Predictive} & \textbf{Aleatoric} & \textbf{Epistemic} & \textbf{Variance} & \textbf{Credible}\\
\hline
\textbf{Conf. Room} & 88.92 \% & 92.56 \%  & 92.59 \% & 91.03 \% & 90.82 \% & \textbf{91.01 \%}\\
\textbf{Copy Room} & 70.82 \% & 72.56 \% & 72.58 \% & 72.37 \% & 72.02 \% & \textbf{74.85 \%}\\
\textbf{Hallway} & 81.13 \% & 83.01 \% & 82.98 \% & 82.98 \% & 83.20 \% & \textbf{93.06 \%}\\
\textbf{Lounge} & 71.77 \% & 73.31 \% & 73.49 \% & 73.09 \% & 73.06 \% & \textbf{77.16 \%}\\
\textbf{Office} & 90.77 \% & \textbf{93.02 \%} & 93.01 \% & 92.28 \% & 92.37 \% & 92.42 \%\\
\textbf{Open Space} & 76.68 \% & 79.45 \% & 79.54 \% & 77.96 \% & 77.77 \% & \textbf{80.38 \%}\\
\textbf{Pantry} & 76.21 \% & 78.51 \% & 78.58 \% & 77.44 \% & 77.24 \% & \textbf{79.20 \%}\\
\hline
\textbf{Assemb. Tact} & 94.21 \% & \textbf{96.63 \%} & 96.64 \% & 95.54 \% & 95.60 \% & 94.99 \%\\
\hline
\end{tabular}
\label{tab:uncertainty-accuracy-bayes}
\end{table} 
The leftmost column contains the accuracy of Bayesian PointNet. The next column contains the accuracy when considering only predictions that have a predictive uncertainty smaller or equal to the mean predictive uncertainty plus two sigma of the predictive uncertainty. The same is displayed in the next columns with respect to aleatoric and epistemic uncertainty as well as the variance of the predictive network outputs. In the last column the predictions for which the \mbox{95~\%-credible interval} of the predicted class overlaps with no other class' \mbox{95~\%-credible interval} are considered certain and are used for prediction. It can be seen that the accuracy increases considerably when only looking at certain predictions with respect to any of the uncertainty measures. Generally, the results for predictive and aleatoric uncertainty as well as the credible interval based method are most promising. This confirms our notion that the predictive uncertainty value is determined mainly by aleatoric uncertainty after thorough network training. The percentage of predictions, which are found to be uncertain, is displayed in Table~\ref{tab:uncertainty-points-bayes}.
\begin{table}
\caption{Percentage of predictions dropped when excluding uncertain predictions in the Bayesian model.}
\centering
%% \tablesize{} %% You can specify the fontsize here, e.g., \tablesize{\footnotesize}. If commented out \small will be used.
\begin{tabular}{ccccccc}
\hline
& \textbf{Bayesian PN} & \textbf{Predictive} & \textbf{Aleatoric} & \textbf{Epistemic} & \textbf{Variance} & \textbf{Credible}\\
\hline
\textbf{Conf. Room} & - & 6.86 \% & 6.90 \% & 5.49 \% & 5.50 \% & 4.07 \% \\
\textbf{Copy Room} & - & 3.25 \% & 3.28 \% & 4.50 \% & 4.81 \% & 9.53 \% \\
\textbf{Hallway} & - & 4.64 \% & 4.65 \% & 5.08 \% & 5.53 \% & 4.54 \% \\
\textbf{Lounge} & - & 3.37 \% & 3.64 \% & 4.89 \% & 5.22 \% & 10.96 \% \\
\textbf{Office} & - & 5.94 \% & 5.94 \% & 4.70 \% & 5.17 \% & 3.73 \% \\
\textbf{Open Space} & - & 6.23 \% & 6.41 \% & 4.97 \% & 5.05 \% & 8.62 \% \\
\textbf{Pantry} & - & 4.75 \% & 4.87 \% & 4.58 \% & 4.78 \% & 7.00 \% \\
\hline
\textbf{Assemb. Tact} & - & 6.55 \% & 6.56 \% & 4.09 \% & 3.70 \% & 1.47 \% \\
\hline
\end{tabular}
\label{tab:uncertainty-points-bayes}
\end{table}
Generally, between about $3$~\% and $11$~\% of the predictions are dropped using the above parameters. The number of dropped predictions decreases when predictions with a higher uncertainty value are considered, e.g. all predictions with uncertainty greater or equal to the mean uncertainty plus three sigma. Generally it can be claimed that the lower the threshold for uncertain predictions, i.e. the more predictions are dropped, the higher the resulting accuracy. Thus, a trade-off between dropping uncertain predictions and segmentation accuracy needs to be found. However, this is largely dependent on the specific use case. Table~\ref{tab:uncertainty-accuracy-dropout} illustrates the results of dropout PointNet for one room of each room type in area 6 of S3DIS data set as well as one tact of car body assembly belonging to the test data set.  
\begin{table}
\caption{Evaluation of different methods for uncertainty estimation and influence on model accuracy in the dropout model.}
\centering
%% \tablesize{} %% You can specify the fontsize here, e.g., \tablesize{\footnotesize}. If commented out \small will be used.
\begin{tabular}{ccccccc}
\hline
& \textbf{Dropout PN} & \textbf{Predictive} & \textbf{Aleatoric} & \textbf{Epistemic} & \textbf{Variance} & \textbf{Credible}\\
\hline
\textbf{Conf. Room} & 87.61 \% & \textbf{90.72 \%} & \textbf{90.72 \%} & 89.66 \% & 89.51 \% & 89.58 \%\\
\textbf{Copy Room} & 71.07 \% & 72.46 \% & 72.55 \% & 72.54 \% & 72.37 \% & \textbf{74.52 \%}\\
\textbf{Hallway} & 81.94 \% & 83.99 \% & 83.97 \% & 83.95 \% & 83.70 \% & \textbf{84.87 \%}\\
\textbf{Lounge} & 70.45 \% & 71.87 \% & 71.89 \% & 72.78 \% & 72.77 \% & \textbf{74.83 \%}\\
\textbf{Office} & 81.06 \% & 82.90 \% & 82.94 \% & 82.46 \% & 82.57 \% & \textbf{83.71 \%}\\
\textbf{Open Space} & 75.97 \% & 78.45 \% & \textbf{78.47 \%} & 77.68 \% & 77.59 \% & 77.96 \%\\
\textbf{Pantry} & 74.44 \% & 76.48 \% & 76.52 \% & 76.87 \% & 76.66 \% & \textbf{77.71 \%}\\
\hline
\textbf{Assemb. Tact} & 94.41 \% & 96.12 \% & \textbf{96.15 \%} & 94.87 \% & 95.00 \% & 94.62 \%\\
\hline
\end{tabular}
\label{tab:uncertainty-accuracy-dropout}
\end{table}
The accuracy of Bayesian PointNet surpasses the accuracy of dropout PointNet for most of the evaluated rooms. In terms of the percentage of predictions dropped the results are similar. Table~\ref{tab:uncertainty-points-dropout} presents the percentage of disregarded predictions compared to the baseline containing all predictions.
\begin{table}
\caption{Percentage of predictions dropped when excluding uncertain predictions in the dropout model.}
\centering
%% \tablesize{} %% You can specify the fontsize here, e.g., \tablesize{\footnotesize}. If commented out \small will be used.
\begin{tabular}{ccccccc}
\hline
& \textbf{Dropout PN} & \textbf{Predictive} & \textbf{Aleatoric} & \textbf{Epistemic} & \textbf{Variance} & \textbf{Credible}\\
\hline
\textbf{Conf. Room} & - & 6.43 \% & 6.38 \% & 5.48 \% & 5.67 \% & 4.30 \% \\
\textbf{Copy Room} & - & 2.96 \% & 3.18 \% & 4.87 \% & 5.43 \% & 8.67 \% \\
\textbf{Hallway} & - & 4.84 \% & 4.89 \% & 5.54 \% & 5.68 \% & 6.96 \% \\
\textbf{Lounge} & - & 2.94 \% & 3.04 \% & 5.22 \% & 5.55 \% & 10.19 \% \\
\textbf{Office} & - & 4.37 \% & 4.45 \% & 4.97 \% & 5.24 \% & 6.71 \% \\
\textbf{Open Space} & - & 6.09 \% & 6.14 \% & 5.92 \% & 5.68 \% & 5.21 \% \\
\textbf{Pantry} & - & 4.37 \% & 4.45 \% & 5.59 \% & 5.84 \% & 7.06 \% \\
\hline
\textbf{Assemb. Tact} & - & 8.44 \% & 8.13 \% & 2.01 \% & 1.69 \% & 9.89 \% \\
\hline
\end{tabular}
\label{tab:uncertainty-points-dropout}
\end{table}
Again, about $2$~\% to $11$~\% of the predictions are dropped by dropout PointNet using the same uncertainty threshold as before. Usually dense point clouds are generated, when building up an environment model of a factory in order to capture as many details as possible. Thus, it is important to keep a high number point wise predictions after uncertainty estimation in order to guarantee high quality when placing object geometries in a simulation model. However, a higher prediction accuracy in the segmentation step also increases the quality of the resulting environment model. As we already discussed, a higher accuracy can be achieved by dropping a larger amount of uncertain predictions. In the case of environment modelling, it is vital to drop as few predictions as possible because otherwise building structures and their exact location that is necessary for model generation can get lost. Generally, we notice that the dropout model is more difficult to train than the Bayesian one, which manifests in a higher epistemic uncertainty in the dropout model. Empirically it is shown that the application of dropout exhibits inferior performance in convolutional architectures~\citep{gal2016dropout}, which could lead to the increased epistemic uncertainty values. Further, the impact of uncertainty on the segmentation performance is more striking in the Bayesian model. However, for applications where one or two percent of accuracy can be sacrificed, the dropout model is a good alternative to the Bayesian model as users can take on a frequentist network and just add dropout during training and test time, without having to define and optimize a distribution over all the network parameters.\\
Overall it can be concluded that Bayesian PointNet has superior performance and dropout PointNet has similar performance to the frequentist model without considering uncertainty information. When only considering certain predictions Bayesian as well as dropout PointNet clearly surpass the performance of the frequentist model. In terms of uncertainty measure the best results are achieved when using the approach using confidence intervals as well as predictive or aleatoric uncertainty. However, the confidence interval based method drops considerably more predictions in some of the examples. Figure~\ref{fig:assembly-tact-inc-unc} displays one tact of car body assembly where certain predictions are displayed in black and uncertain predictions are displayed in red. The applied model corresponds to Bayesian PointNet with the credible interval based uncertainty measure.
It shows that the network is certain about the majority of its predictions. Uncertain predictions are concentrated at the ceiling, wall and columns, which is in line with our expectations of Section~\ref{subsec:automotive-factory-dataset}. As the ceiling but especially the walls and columns are hung with clutter objects, this leads to uncertain predictions as the point clouds of these classes are incomplete due to holes. Further, Figure~\ref{fig:boxplot} shows the boxplots of the predictive softmax outputs in the Bayesian model for a correct and a wrong prediction of two single points in the automotive factory data set. In Figure~\ref{fig:boxplot}~(a) it can be seen that the network is certain about its correct prediction, i.e. all the predictive softmax output values of the correct class are close to one, while the network outputs for all other classes are close to zero. In the case of a wrong prediction, see Figure~\ref{fig:boxplot}~(b), the boxes of the true and predicted label overlap indicating an uncertain prediction. The white box corresponds to the correct class and the shaded box corresponds to the wrongly predicted class.

\section{Discussion and Conclusion}
\label{sec:discussion-and-conclusion}
The use of Bayesian neural networks instead of frequentist ones allows the quantification of network uncertainty. On the one hand, this leads to more robust and accurate models. On the other hand, in safety critical application uncertain prediction can be identified and treated with special care. For the use case of point cloud segmentation in modelling production sites there are hardly any safety critical issues, however, an increased model accuracy leads to more accurate reconstructions of the real-world production system in a simulation engine or in CAD software. A first methodology for systematic data collection and processing in a large-scale industrial environment was presented in~\citep{petschnigg2020point}. In future work this will be extended to a more thorough workflow including technology specifications and a mathematical concept of how to place the segmented objects in a simulation model.\\
In summary, in this work we present a novel Bayesian neural network that is capable of 3D deep semantic segmentation of raw point clouds. Further, it allows the estimation of uncertainty in the network predictions. Additionally, a network using dropout training to approximate Bayesian variational inference in Gaussian processes is described. We compare the uncertainty information gained by the Bayesian and the dropout model. Both the Bayesian and the dropout model effectively increase the network performance during test time compared to the frequentist framework when taking into account the information gained by network uncertainty. The dropout model shows on-par performance to frequentist PointNet without taking into account network uncertainty. Bayesian PointNet is more robust against overfitting than the frequentist one and achieves higher test accuracy even without considering uncertainty information. Bayesian segmentation allows to work with fewer example data, due to prior information acting like additional observations, while computational complexity basically stays the same. All of the proposed networks are embedded in an industrial prototype that aims at generating static simulation models out of a raw point cloud only. The complete system will be evaluated in our next work.

\section{Conflict of interest}
The authors declare that they have no conflict of interest.

\bibliographystyle{spbasic}      % basic style, author-year citations
\bibliography{references}   % name your BibTeX data base

%\begin{thebibliography}{}
% etc
%\end{thebibliography}

\end{document}